\documentclass[11pt,a4paper]{article}
\usepackage[hyperref]{naaclhlt2019}
\usepackage{times}
\usepackage{latexsym}

\usepackage{url}

\usepackage{amsmath}
\usepackage{amssymb}
\usepackage{multirow}
\usepackage{array}
\usepackage{graphicx}
\usepackage{algorithm}
\usepackage{algorithmic}
\usepackage{enumitem}
\usepackage{comment}

\aclfinalcopy 

\title{An Integrated Approach for Keyphrase Generation via Exploring the Power of Retrieval and Extraction}

\author{Wang Chen\textsuperscript{\rm 1}, 
Hou Pong Chan\textsuperscript{\rm 1}, 
Piji Li\textsuperscript{\rm 3},
Lidong Bing\textsuperscript{\rm 2},
Irwin King\textsuperscript{\rm 1} \\
\textsuperscript{\rm 1}The Chinese University of Hong Kong, Shatin, N.T., Hong Kong \\
\textsuperscript{\rm 2}R\&D Center Singapore, Machine Intelligence Technology, Alibaba DAMO Academy\\
\textsuperscript{\rm 3}Tencent AI Lab\\
\textsuperscript{\rm 1}\{wchen, hpchan, king\}@cse.cuhk.edu.hk, \textsuperscript{\rm 3}pijili@tencent.com, 
\textsuperscript{\rm 2}l.bing@alibaba-inc.com
}

\date{}

\begin{document}
\maketitle
\begin{abstract}
In this paper, we present a novel integrated approach for keyphrase generation (KG). Unlike previous works which are purely extractive or generative, we first propose a new multi-task learning framework that jointly learns an extractive model and a generative model. Besides extracting keyphrases, the output of the extractive model is also employed to rectify the copy probability distribution of the generative model, such that the generative model can better identify important contents from the given document. Moreover, we retrieve similar documents with the given document from training data and use their associated keyphrases as external knowledge for the generative model to produce more accurate keyphrases. For further exploiting the power of extraction and retrieval, we propose a neural-based merging module to combine and re-rank the predicted keyphrases from the enhanced generative model, the extractive model, and the retrieved keyphrases.
Experiments on the five KG benchmarks demonstrate that our integrated approach outperforms the state-of-the-art methods.
\end{abstract}

\section{Introduction}
\begin{figure}[t]
\centering
\includegraphics[width=\columnwidth]{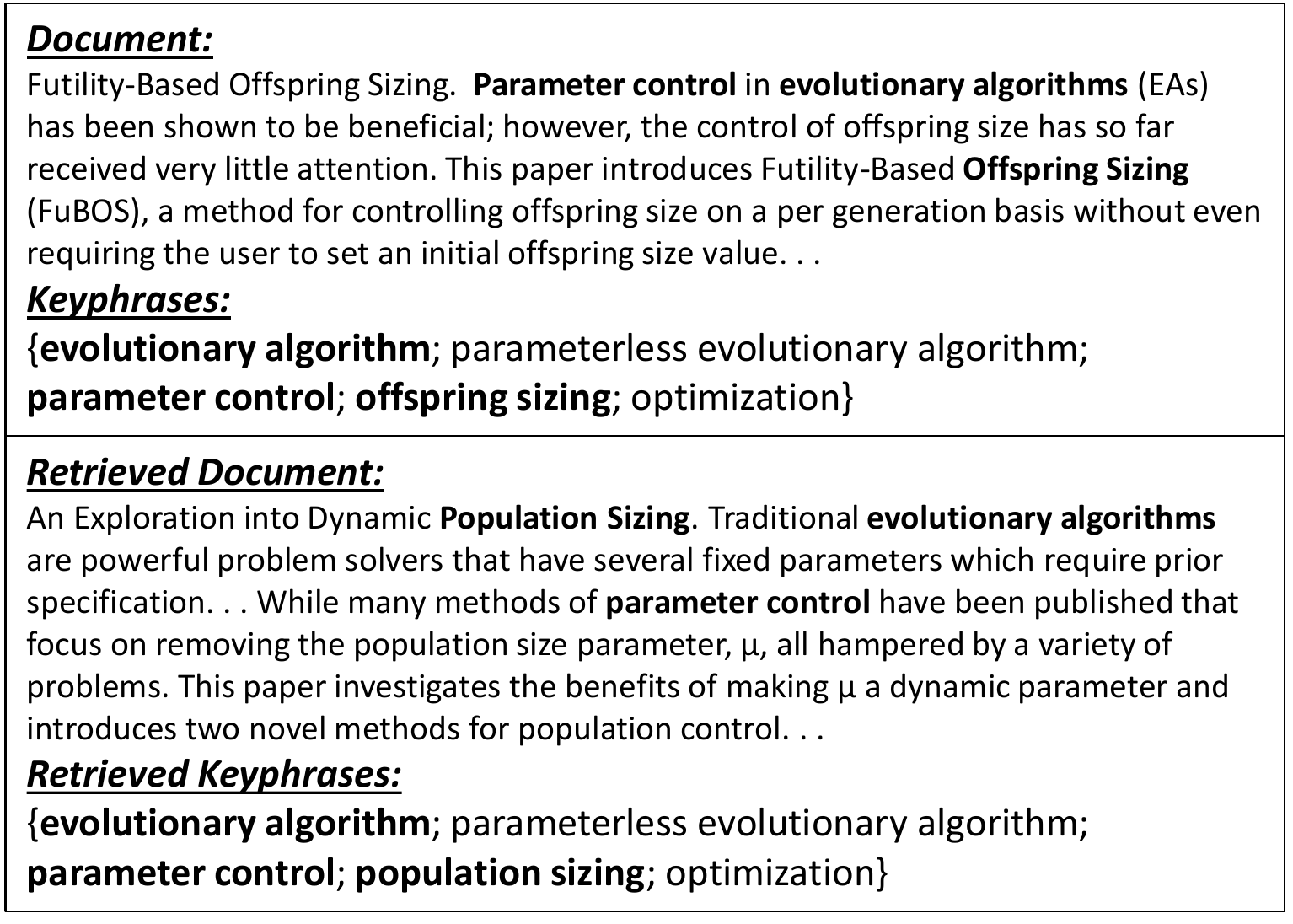}
\caption{An example of keyphrase generation and retrieval. The present keyphrases are bold.}
\label{figure: keyphrase generation example}
\end{figure}

Keyphrases are short text pieces that can quickly express the key ideas of a given document. The keyphrase generation task aims at automatically generating a set of keyphrases given a document. As shown in the upper part of Figure~\ref{figure: keyphrase generation example}, the input is a document and the output is a set of keyphrases. Due to the concise and precise expression, keyphrases are beneficial to extensive downstream applications such as text summarization~\cite{zhang2004world_summarize_app, wang2013domain_summarize_app}, sentiment analysis~\cite{wilson2005recognizing_senti_app,berend2011opinion_opin_app}, and document clustering~\cite{hulth2006study_doc_clus_app,hammouda2005corephrase_doc_clus_app}.

Existing methods on keyphrase generation can be divided into two categories: \textit{extractive} and \textit{generative}.
Extractive methods~\cite{medelyan2009human_maui, mihalcea2004textrank, zhang2016keyphrase_seqlabel_twitter, luan2017scientific_seqlabel} identify present keyphrases that appear in the source text like ``\mbox{parameter} control'' in Figure~\ref{figure: keyphrase generation example}. Although extractive methods are simple to implement, they cannot predict absent keyphrases which are not in the document like ``optimization'' in Figure~\ref{figure: keyphrase generation example}. 
Generative methods~\cite{Meng2017dkg, Chen2018corr_dkg,Ye2018semi_dkg, yuan2018generating_diverse} adopt the well-known encoder-decoder generative model~\cite{luong2015effective, DBLP:conf/iclr/BahdanauCB14} with copy mechanism~\cite{Gu2016copy,see2017get} to produce keyphrases. In a generative model, the decoder generates keyphrases word by word through either selecting from a predefined vocabulary according to a language model or copying from the source text according to the copy probability distribution computed by a copy mechanism. Thus, these generative methods are capable of generating both present and absent keyphrases. 

From a high-level perspective, extractive methods directly locate essential phrases in the document while generative models try to understand the document first and then produce keyphrases. To the best of our knowledge, these two kinds of methods have been developing independently without any combinations among them. 

However, when human annotators are asked to assign keyphrases to a document, they usually first obtain a global sense about which parts of the document are important and then write down the keyphrases word by word based on a more detailed understanding. To achieve such a goal, we propose a multi-task learning framework to take advantage of both extractive and generative models. 
For keyphrase extraction, we adopt a neural sequence labeling model to output the likelihood of each word in the source text to be a keyphrase word (or the importance score of each word).
These importance scores are then employed to rectify the copy probability distribution of the generative model. 
Since the extractive model is explicitly trained to identify keyphrases from the source text, its importance scores can help the copy mechanism to identify important source text words more accurately. 
Different from the copy probability distribution which is dynamic at each generation step, these importance scores are static. Therefore, they can provide a global sense about which parts of the document are important. 
In addition, these scores are also utilized to extract present keyphrases which will be exploited by the merging module.

Moreover, human annotators can also incorporate relevant external knowledge like the keyphrases of similar documents that they read before to assign more appropriate keyphrases. Correspondingly, to incorporate external knowledge, we propose a \textit{retriever} to retrieve similar documents of the given document from training data. For instance, as shown in Figure~\ref{figure: keyphrase generation example}, we retrieve a document from the \textbf{KP20k} training dataset that has the highest similarity with the upper document. The retrieved document is assigned with almost the same keyphrases as the upper document. Therefore, keyphrases from similar documents (i.e., retrieved keyphrases) can give useful knowledge to guide the generation of keyphrases for the given document. 
More concretely, we encode the retrieved keyphrases as vector representations and use them as an external memory for the decoder of the generative model in our multi-task learning framework. Besides providing external knowledge, the retrieved keyphrases themselves are regarded as a kind of keyphrase prediction and can be utilized by the merging module.

Finally, to imitate the integrated keyphrase assignment process of humans more comprehensively, we further exploit the extractive model and the retrieved keyphrases by proposing a merging module. This merging module collects and re-ranks the predictions from our aforementioned components. First, keyphrase candidates are collected from three different sources: (1) keyphrases generated by the enhanced generative model; (2) keyphrases extracted by the extractive model; and (3) the retrieved keyphrases. Then, we design a neural-based merging algorithm to merge and re-rank all the keyphrase candidates, and consequently return the top-ranked candidates as our final keyphrases. 
 
We extensively evaluate the performance of our proposed approach on five popular benchmarks. 
Experimental results demonstrate the effectiveness of the extractive model and the retrieved keyphrases in our multi-task learning framework.
Furthermore, after introducing the merging module, our integrated approach consistently outperforms all the baselines and becomes the new state-of-the-art approach for keyphrase generation.

In summary, our main contributions include: (1) a new multi-task learning framework that leverages an extractive model and external knowledge to improve keyphrase generation; (2) a novel neural-based merging module that combines the predicted keyphrases from extractive, generative, and retrieval methods to further improve the performance; and (3) the new state-of-the-art performance on five real-world benchmarks.

\section{Related Work}
\subsection{Automatic Keyphrase Extraction}
Keyphrase extraction focuses on predicting the keyphrases that are present in the source text. 
Existing methods can mainly be categorized into two-step extraction approaches and sequence labeling models. Two-step extraction approaches first identify a set of candidate phrases from the document using different heuristics, such as the phrases that match specific part-of-speech (POS) tags~\cite{DBLP:conf/conll/LiuCZS11,DBLP:conf/iconip/WangZH16,DBLP:conf/ausai/LeNS16}. Then, they learn a score for each candidate and select the top-ranked candidates as predicted keyphrases. The scores can be learned by either supervised methods with hand-crafted textual features~\cite{medelyan2009human_maui, witten2005kea,Nguyen2007NUS,DBLP:conf/ijcai/FrankPWGN99,Hulth2003inspec} or unsupervised graph ranking methods~\cite{mihalcea2004textrank,DBLP:conf/www/GrinevaGL09, DBLP:conf/aaai/WanX08}. Sequence labeling models are built on a recurrent neural network to sequentially go through a source text and learn the likelihood of each word in the source text to be a keyphrase word~\cite{zhang2016keyphrase_seqlabel_twitter,luan2017scientific_seqlabel,gollapalli2017incorporating_seqlabel}. In contrast to these extractive methods, our approach can generate both absent and present keyphrases.

\subsection{Automatic Keyphrase Generation}
Keyphrase generation aims at predicting both present and absent keyphrases for a source text. \citet{Meng2017dkg} proposed CopyRNN, which is built on the attentional encoder-decoder model~\cite{DBLP:conf/iclr/BahdanauCB14} with copy mechanism~\cite{Gu2016copy} to generate keyphrases. 
CorrRNN~\cite{Chen2018corr_dkg}, an extension of CopyRNN, was proposed to model the correlations among keyphrases. This model utilizes hidden states and attention vectors of previously generated keyphrases to avoid generating repetitive keyphrases. 
The title information of the source text was explicitly exploited by \newcite{Ye2018semi_dkg} and \newcite{Chen2018TG_net} to further improve the performance. \citet{Ye2018semi_dkg} first considered a semi-supervised setting for keyphrase generation.
In contrast, inspired by \newcite{hsu2018unified_inconsistency} and \newcite{cao2018re3sum}, we enhance existing generative methods by adopting an extractive model to assist the copy mechanism and exploiting external knowledge from retrieved keyphrases to help the generation. 
Furthermore, we also design a merging module to combine the predictions from different components.

\section{Our Methodology}
\begin{figure}[t]
\centering
\includegraphics[width=\columnwidth]{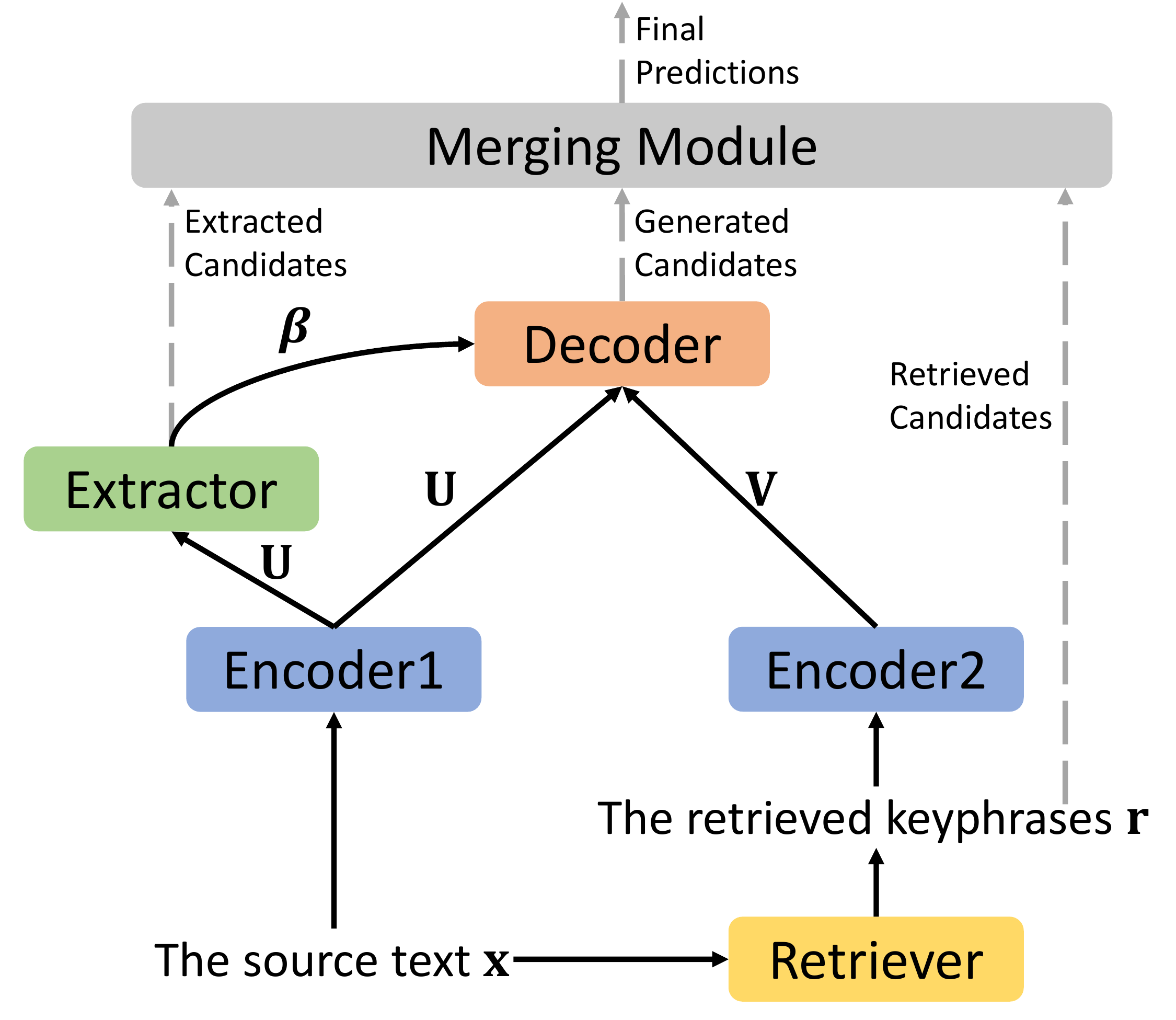}
\caption{Our integrated framework. The ``Encoder1'' and the ``Extractor'' compose our extractive model. Our generative model mainly includes the ``Encoder1'', the ``Encoder2'', and the ``Decoder''.}
\label{figure: KG_KE_RK_merge model}
\end{figure}

As shown in Figure~\ref{figure: KG_KE_RK_merge model}, our integrated framework consists of a retriever, two encoders, an extractor, a decoder, and a merging module. Given a document $\mathbf{x}$, the retriever returns the keyphrases $\mathbf{r}$ retrieved from the training corpus. In addition to acting as keyphrase candidates, these retrieved keyphrases are also exploited to provide external guidance for the decoder. Then keyphrase extraction and generation are jointly conducted by the extractor and the decoder through sharing an encoder. Besides extracting keyphrase candidates, the importance scores of the source text words, $\boldsymbol{\beta}$, predicted by the extractor are also employed to rescale the original copy probability distribution of the decoder. Thus, they can help the copy mechanism to detect important words more accurately. Finally, the merging module merges the candidates from three different sources (i.e., the retrieved, extracted, and generated candidates) and output the final predictions. 

\subsection{Retriever} \label{retrieval_module}
Given a document $\mathbf{x}$, the retriever module retrieves top $K$ (document, keyphrases) pairs from the training corpus. The retrieval is based on the Jaccard similarities of the non-stop-word sets between $\mathbf{x}$ and the corpus documents. After that, the keyphrases of the top $K$ pairs are returned and used in the later modules in two ways. First, these retrieved keyphrases are regarded as the keyphrase candidates of $\mathbf{x}$ and directly fed into the final merging module. In addition, these keyphrases are concatenated together as a guidance input $\mathbf{r}$ for the decoder to provide useful external knowledge for the generation process. A separator token is inserted among keyphrases when concatenating them together.

\subsection{Joint Extraction and Generation}
We propose a multi-task learning framework which simultaneously learns to extract keyphrases from the source text and generate keyphrases word by word with external knowledge. Before describing in detail, we first define the tasks of the extraction and the generation.

\subsubsection{Problem Definition}
The inputs of the multi-task learning framework are the source text $\mathbf{x}$ and the concatenated \mbox{retrieved} keyphrases $\mathbf{r}$. Both $\mathbf{x}$ and $\mathbf{r}$ are a sequence of tokens (i.e., $\mathbf{x}=[x_1, ..., x_{L_\mathbf{x}}]$, $\mathbf{r}=[r_1, ..., r_{L_\mathbf{r}}]$), where $L_\mathbf{x}$ and $L_\mathbf{r}$ are the length of $\mathbf{x}$ and $\mathbf{r}$ respectively. 
The output of the extractor is a sequence of importance scores $\boldsymbol{\beta}=[\beta_1, ..., \beta_{L_{\mathbf{x}}}]$, where $\beta_{i}$ is the probability of the $i$-th source word of being a keyphrase word. 
The output of the generator is a set of keyphrases $\mathcal{Y}=\{\mathbf{y}^i\}_{i=1,..,N}$, where $N$ is the keyphrase number of $\mathbf{x}$ and $\mathbf{y}^i=[y^i_1, ..., y^i_{L_{\mathbf{y}^i}}]$ is a token sequence with length $L_{\mathbf{y}^i}$.

To fit the encoder-decoder framework, $N$ tuples $\{(\mathbf{x}, \mathbf{r}, \boldsymbol{\beta}^*, (\mathbf{y}^i)^*)\}_{i=1,...,N}$ are split during training, where $\boldsymbol{\beta}^*$ and $(\mathbf{y}^i)^*$ are the gold binary importance scores and one of the gold keyphrases of $\mathbf{x}$ correspondingly. For simplicity, we adopt $(\mathbf{x}, \mathbf{r}, \boldsymbol{\beta}^*, \mathbf{y}^{*})$ to represent such a tuple.

\subsubsection{Encoders}
Two encoders are employed in our multi-task learning framework. One is for the source text encoding (i.e., ``Encoder1'' in Figure~\ref{figure: KG_KE_RK_merge model}) and the other is for retrieved keyphrases encoding (i.e., ``Encoder2'' in Figure~\ref{figure: KG_KE_RK_merge model}). Both of them employ a bidirectional GRU~\cite{cho2014gru} layer to obtain a context-aware representation of each word:
\begin{align}
\mathbf{u}_i &= \text{BiGRU}_{1}(\mathbf{x}_i, \overrightarrow{\mathbf{u}}_{i-1}, \overleftarrow{\mathbf{u}}_{i+1}), \\
\mathbf{v}_j &= \text{BiGRU}_{2}(\mathbf{r}_j, \overrightarrow{\mathbf{v}}_{j-1}, \overleftarrow{\mathbf{v}}_{j+1}),
\end{align}
where $i=1,2,...,L_{\mathbf{x}}$ and $j=1,2,...,L_{\mathbf{r}}$. $\mathbf{x}_i$ and $\mathbf{r}_j$ are the $d_e$-dimensional embedding vectors of the $i$-th source text word $x_i$ and $j$-th retrieved keyphrases word $r_j$ respectively. $\mathbf{u}_i=[\overrightarrow{\mathbf{u}}_i;\overleftarrow{\mathbf{u}}_i] \in \mathbb{R}^{d}$ and $\mathbf{v}_j=[\overrightarrow{\mathbf{v}}_j;\overleftarrow{\mathbf{v}}_j] \in \mathbb{R}^{d}$ are regarded as the corresponding context-aware representations, where $d$ is the hidden size of the biderectional GRU layer. Finally, we obtain the internal memory bank $\mathbf{U}=[\mathbf{u}_1, ..., \mathbf{u}_{L_{\mathbf{x}}}]$ for later extraction and generation, and the external memory bank $\mathbf{V}=[\mathbf{v}_1, ..., \mathbf{v}_{L_{\mathbf{r}}}]$ for later generation.

\subsubsection{Extractor}
Based on the internal memory bank, we use the following sequence identifier as our extractor to identify whether the word is a keyphrase word in the source text. We denote the importance score \mbox{$P(\beta_j=1 | \mathbf{u}_j, \mathbf{s}_j, \mathbf{d})$} as $\beta_j$ for simplicity:
\begin{align}
\begin{split}
    \beta_j = \text{sigmoid}(& \mathbf{W}_c \mathbf{u}_j + \mathbf{u}_j^T \mathbf{W}_s \mathbf{d}\\
    &- \mathbf{u}_j^T \mathbf{W}_n \text{tanh}(\mathbf{s}_j) + b),
\end{split}
\end{align}
where $\mathbf{d} = \text{tanh}(\mathbf{W}_d[\overrightarrow{\mathbf{u}}_{L_{\mathbf{x}}};\overleftarrow{\mathbf{u}}_1] + \mathbf{b})$ is the global document representation and $\mathbf{s}_j = \sum_{i=1}^{j-1} \mathbf{u}_i \beta_i$ is current summary representation. $\mathbf{W}_c, \mathbf{W}_s$, and $\mathbf{W}_n$ are the content, salience and novelty weights respectively. Although this extractor is inspired by \newcite{nallapati2017summarunner}, our extractor identifies important words instead of sentences within the source text.

\subsubsection{Decoder}
In addition to the internal memory bank $[\mathbf{u}_1, ..., \mathbf{u}_{L_{\mathbf{x}}}]$, our decoder employs the external memory bank $[\mathbf{v}_1, ..., \mathbf{v}_{L_{\mathbf{r}}}]$ to provide external guidance for the generation process.
We exploit a decoder equipped with attention and copy mechanisms~\cite{luong2015effective,see2017get} to generate keyphrases.
This decoder mainly consists of a forward GRU layer:
\begin{align}
    \mathbf{h}_t & = \overrightarrow{\text{GRU}}([\mathbf{e}_{t-1};\tilde{\mathbf{h}}_{t-1}], \mathbf{h}_{t-1}), \\
    \mathbf{c}_t^{in} & = \text{attn}(\mathbf{h}_t, [\mathbf{u}_1, ..., \mathbf{u}_{L_{\mathbf{x}}}], \mathbf{W}_{in}), \label{eq:internal_contextual_vec} \\
    \mathbf{c}_t^{ex} & = \text{attn}(\mathbf{h}_t, [\mathbf{v}_1, ..., \mathbf{v}_{L_{\mathbf{r}}}], \mathbf{W}_{ex}), \label{eq:external_contextual_vec} \\
    \tilde{\mathbf{h}}_t & = \text{tanh}(\mathbf{W}_1[\mathbf{c}_t^{in}; \mathbf{c}_t^{ex}];\mathbf{h}_t),
\end{align}
where $\mathbf{e}_{t-1}$ is the embedding vector of the $(t-1)$-th predicted word. The ``attn'' operation in Eq.~(\ref{eq:internal_contextual_vec}) is defined as $\mathbf{c}_t^{in} =  \sum_{i=1}^{L_{\mathbf{x}}} \alpha_{t,i}^{in} \mathbf{u}_i$, where $\alpha_{t,i}^{in} = \exp(s_{t,i}) \slash \sum_{j=1}^{L_{\mathbf{x}}} \exp(s_{t,j})$ and $s_{t,i} = (\mathbf{h}_t)^T \mathbf{W}_{in} \mathbf{u}_i$. Similarly, we can obtain the external aggregated vector $\mathbf{c}_t^{ex}$.

Then, the final predicted probability distribution at the current time step is:
\begin{align}
    P(y_t) = (1-g_t) P_v(y_t) + g_t P_c(y_t) \text{,}
\end{align}
where $g_t = \sigma(\mathbf{w}^T_g\tilde{\mathbf{h}}_{t} + b_g) \in \mathbb{R}$ is the soft switch between generating from the predefined vocabulary $V$ and copying from $X$ that are all words appearing in the source text. $P_v(y_t)=\text{softmax}(\mathbf{W}_2\tilde{\mathbf{h}}_{t} + \mathbf{b}_v) \in \mathbb{R}^{|V|}$ is the generating probability distribution over $V$ and $P_c(y_t)=\sum_{i:x_i=y_t}\alpha_{t,i}^{c} \in \mathbb{R}^{|X|}$ is the copying probability distribution over $X$. Previous work either directly uses the internal attention scores as the copy probabilities (i.e., $\alpha_{t,i}^c = \alpha_{t,i}^{in}$) or employs extra neural network layers to calculate new copy scores. But we employ the rescaled internal attention scores $\alpha_{t}^{in}$ by the importance scores $[\beta_1, ..., \beta_{L_{\mathbf{x}}}]$ from the extractor as the final copy probabilities:
\begin{align}
\alpha_{t,i}^{c} = \frac{\alpha_{t,i}^{in} * \beta_i}{\sum_{j=1}^{L_{\mathbf{x}}} \alpha_{t,j}^{in} * \beta_j}.
\end{align}
The purpose of this rescaling is to provide extra guidance that which words within the source text are important and thus should obtain more attention when copying.

\subsubsection{Joint Training}
Finally, the summation of the following extraction loss and generation loss is used to train the whole joint framework in an end-to-end way.

\textbf{Extraction Loss.} We choose the source text words appearing in the assigned keyphrases as the gold important words and use the weighted cross-entropy loss for the extraction training i.e., $\mathcal{L}_e = -\frac{1}{L_\mathbf{x}} \sum_{j=1}^{L_\mathbf{x}} w \beta^*_j \text{log} \beta_j + (1 - \beta^*_j) \text{log} (1 - \beta_j)$, where $\beta^*_j \in \{0,1\}$ is the ground-truth label for the $j$-th word and $w$ is the loss weight for the positive training samples.

\textbf{Generation Loss.} The negative log likelihood loss is utilized for the generation training i.e., $\mathcal{L}_g = -\sum_{t=1}^{L_{\mathbf{y}^*}} \text{log} P(y_t^*| \mathbf{y}_{t-1}, \mathbf{x}, \mathbf{r})$, where $\mathbf{y}_t=[y_1, ..., y_{t-1}]$ is the previously predicted word sequence, $L_{\mathbf{y}^*}$ is the length of target keyphrase $\mathbf{y}^*$, and $y_t^*$ is the $t$-th target word in $\mathbf{y}^*$.

\subsection{Merging Module}
In this module, the retrieved, extracted and generated keyphrases are collected and then merged to produce the final keyphrase predictions.

\subsubsection{Keyphrase Candidate Collection}
\textbf{Retrieved Candidate Collection.} The retrieved keyphrases from the retriever are regarded as the retrieved candidates. Each retrieved candidate ($rk$) obtains a retrieval score ($rs$) that is the Jaccard similarity between the corresponding document and $\mathbf{x}$. The duplicates with lower retrieval scores are removed. Finally, we get $N_{\mathbf{rk}}$ retrieved keyphrase candidates $\mathbf{rk}=[rk_1,\dots,rk_{N_{\mathbf{rk}}}]$ and their retrieval scores $\mathbf{rs}=[rs_1,\dots,rs_{N_{\mathbf{rk}}}]$.

\textbf{Extracted Candidate Collection.} The extracted keyphrase candidates are from the extractor. We select the word $x_j$ as a keyword if its importance score $\beta_j$ is larger or equal than a threshold $\epsilon$ (i.e., $\beta_j \geq \epsilon$). The adjacent keywords compound a keyphrase candidate. If no other adjacent keywords, the keyword itself becomes a single-word keyphrase candidate. Each extracted keyphrase candidate ($ek$) is accompanied by an extraction score ($es$) that is the mean of the importance scores of the words within this candidate. Similarly, duplicates with lower extraction scores are removed. Consequently, we obtain $N_{\mathbf{ek}}$ extracted keyphrase candidates $\mathbf{ek}=[ek_1,\dots,ek_{N_{\mathbf{ek}}}]$ and the corresponding extraction scores $\mathbf{es}=[es_1,\dots,es_{N_{\mathbf{ek}}}]$.

\textbf{Generated Candidate Collection.} The generated keyphrase candidates directly come from the beam search process of the decoder. Each generated phrase is a keyphrase candidate. The beam search score of the generated candidate ($gk$) represents its generation score ($gs$). Duplicates with lower generation scores are removed. Then, we get $N_{\mathbf{gk}}$ generated candidates $\mathbf{gk}=[gk_1,\dots,gk_{N_{\mathbf{gk}}}]$ and their generation scores $\mathbf{gs}=[gs_1,\dots,gs_{N_{\mathbf{gk}}}]$.

\subsubsection{Merging}
In addition to the original importance scores (i.e., $\mathbf{rs, es, gs}$), we also employ an auxiliary \textit{scorer} to assign an auxiliary importance score to each keyphrase candidate. Given a document-candidate pair ($\mathbf{x}$, candidate), the scorer should output the probability that the candidate is one of the keyphrases of $\mathbf{x}$. That means the scorer should determine the relationship between the given document $\mathbf{x}$ and the candidate, which is similar to a natural language inference (NLI) problem. Therefore, we adapt the most popular NLI model~\cite{Parikh2016model_nli} as our scorer. Different from typical natural language inference which is a multi-class classification problem, we use a binary classification setting to train the scorer. Besides, we learn the word embeddings and use two bi-directional GRU to obtain the input representations. The positive samples are the ground-truth keyphrases. The negative samples come from either the phrases in the document or the retrieved candidates. Notably, the ground-truth keyphrases are filtered when selecting negative samples. Consequently, a cross-entropy loss is utilized to train the scorer. Finally, the trained scorer is used to help the merging process as shown in Algorithm~\ref{alg:merging}. The $\frac{u_{\mathbf{gs}}}{u_{\mathbf{rs}}}$ and $\frac{u_{\mathbf{gs}}}{u_{\mathbf{es}}}$ factors are used to enforce the average of $\mathbf{rs}$ and $\mathbf{es}$ to be the same with the average of $\mathbf{gs}$ and thus these three scores become comparable. 

\begin{algorithm}[t]
\small
\caption{Merging Algorithm}
\label{alg:merging}
\begin{algorithmic}[1]
\REQUIRE The retrieved, extracted and generated candidates $\mathbf{rk, ek, gk}$. The retrieval, extraction and generation scores $\mathbf{rs, es, gs}$; The average of each kind of score: $u_{\mathbf{rs}}, u_{\mathbf{es}}, u_{\mathbf{gs}}$; The trained $scorer$; The document $\mathbf{x}$.
\STATE Adjust $\mathbf{gs}$: $gs_i = gs_i \times scorer(\mathbf{x},gk_i)$ where $i=1, \dots, N_{\mathbf{gk}}$.
\STATE Adjust $\mathbf{rs}$: $rs_i = rs_i \times \frac{u_{\mathbf{gs}}}{u_{\mathbf{rs}}} \times scorer(\mathbf{x},rk_i)$ where $i=1, \dots, N_{\mathbf{rk}}$. \label{adjust_rs}
\STATE Adjust $\mathbf{es}$: $es_i = es_i \times \frac{u_{\mathbf{gs}}}{u_{\mathbf{es}}} \times scorer(\mathbf{x},ek_i)$ where $i=1, \dots, N_{\mathbf{ek}}$. \label{adjust_es}
\STATE Merge $\mathbf{rk, ek, gk}$: the final importance score of a candidate is the summation of its adjusted retrieval, extraction and generation scores. If not in $\mathbf{rk}$, $\mathbf{ek}$ or $\mathbf{gk}$, the corresponding scores are set to 0.
\STATE Sort all the candidates based on the final importance scores and then output the final predictions.
\end{algorithmic}
\end{algorithm}

\section{Experiment Settings}
\subsection{Datasets}
Similar to~\newcite{Meng2017dkg}, we use \textbf{KP20k} dataset~\cite{Meng2017dkg} to train our models. The released dataset contains 530,809 articles for training, 20,000 for validation, and the other 20,000 for testing. However, there exist duplicates in the \textbf{KP20k} training dataset with itself, the \textbf{KP20k} validation dataset, the \textbf{KP20k} testing dataset, and other four popular testing datasets (i.e., \textbf{Inspec}~\cite{Hulth2003inspec}, \textbf{Krapivin}~\cite{Krapivin2009LargeDF}, \textbf{NUS}~\cite{Nguyen2007NUS}, and \textbf{SemEval}~\cite{Kim2010SemEval}). After removing these duplicates, we maintain 509,818 articles in the training dataset. As for testing, following~\newcite{Meng2017dkg}, we employ five popular testing datasets from scientific publications as our testbeds for the baselines and our methods, which include \textbf{Inspec}, \textbf{Krapivin}, \textbf{NUS}, \textbf{SemEval}, and \textbf{KP20k}.

\subsection{Baseline Models and Evaluation Metrics}
For a comprehensive evaluation, we compare our methods with the traditional extractive baselines and the state-of-the-art generative methods. The extractive baselines include two unsupervised methods (i.e., TF-IDF and TextRank~\cite{mihalcea2004textrank}) and one supervised method Maui~\cite{medelyan2009human_maui}. The generative baselines consist of CopyRNN~\cite{Meng2017dkg} and CorrRNN~\cite{Chen2018corr_dkg}. We also conduct several ablation studies as follows:
\begin{itemize}[leftmargin=*]
    \item \textbf{KG-KE.} The joint extraction and generation model without using the retrieved keyphrases and merging process.
    \item \textbf{KG-KR.} The encoder-decoder generative model with retrieved keyphrases as external knowledge, but without combining with the extractive model and using the merging process.
    \item \textbf{KG-KE-KR.} The joint extraction and generation model with the retrieved keyphrases without using the merging process.
\end{itemize}
All the above ablation models directly use the generated candidates as the final predictions. We denote our final integrated method which combines all the proposed modules as \textbf{KG-KE-KR-M}.

Similar to CopyRNN and CorrRNN, we adopt macro-averaged \textit{recall} (R) and \textit{F-measure} ($\text{F}_1$) as our evaluation metrics. In addition, we also apply Porter Stemmer before determining whether two keyphrases are identical. Duplications are removed after stemming.

\begin{table*}
\centering
\small
\begin{tabular}{>{\centering\arraybackslash}p{2.0cm}| 
>{\centering\arraybackslash}p{0.85cm} >{\centering\arraybackslash}p{0.95cm}| 
>{\centering\arraybackslash}p{0.85cm} >{\centering\arraybackslash}p{0.95cm}| 
>{\centering\arraybackslash}p{0.85cm} >{\centering\arraybackslash}p{0.95cm}| 
>{\centering\arraybackslash}p{0.85cm} >{\centering\arraybackslash}p{0.95cm}| 
>{\centering\arraybackslash}p{0.85cm} >{\centering\arraybackslash}p{0.95cm}}
 
  \hline
  \hline
 \multirow{2}{*}{\textbf{Model}} & \multicolumn{2}{c|}{\textbf{Inspec}} & \multicolumn{2}{c|}{\textbf{Krapivin}} & \multicolumn{2}{c|}{\textbf{NUS}} & \multicolumn{2}{c|}{\textbf{SemEval}} & \multicolumn{2}{c}{\textbf{KP20k}}
  \\
  & $\text{F}_1$@5 & $\text{F}_1$@10 & $\text{F}_1$@5 & $\text{F}_1$@10 & $\text{F}_1$@5 & $\text{F}_1$@10 & $\text{F}_1$@5 & $\text{F}_1$@10 & $\text{F}_1$@5 & $\text{F}_1$@10
  \\
  \hline
  \hline
  TF-IDF & 0.188 & 0.269 & 0.092 & 0.120 & 0.103 & 0.142 & 0.076 & 0.135 & 0.087 & 0.113
  \\
  TextRank & 0.194 & 0.244 & 0.142 & 0.128 & 0.147 & 0.153 & 0.107 & 0.130 & 0.151 & 0.132
  \\
  Maui & 0.037 & 0.032 & 0.196 & 0.181 & 0.205 & 0.234 & 0.032 & 0.036 & 0.223 & 0.204
  \\
  \hline
  CorrRNN\textsuperscript{*} 
  & 0.229\textsubscript{7} & 0.248\textsubscript{9} 
  & 0.255\textsubscript{2} & 0.238\textsubscript{4} 
  & 0.273\textsubscript{5} & 0.265\textsubscript{4} 
  & 0.197\textsubscript{3} & 0.221\textsubscript{5} 
  & 0.291\textsubscript{2} & 0.264\textsubscript{2}
  \\
  CopyRNN\textsuperscript{*} 
  & 0.251\textsubscript{7} & 0.279\textsubscript{3} 
  & \underline{0.268\textsubscript{4}} & 0.243\textsubscript{1} 
  & 0.275\textsubscript{2} & 0.268\textsubscript{2} 
  & 0.190\textsubscript{6} & 0.214\textsubscript{5} 
  & 0.306\textsubscript{1} & 0.273\textsubscript{0}
  \\
  \hline
  KG-KE 
  & \underline{0.254\textsubscript{4}} & \underline{0.281\textsubscript{2}} 
  & 0.265\textsubscript{3} & 0.240\textsubscript{1} 
  & 0.278\textsubscript{4} & 0.273\textsubscript{1} 
  & \textbf{0.207\textsubscript{4}} & \textbf{0.227\textsubscript{7}} 
  & 0.307\textsubscript{0} & 0.274\textsubscript{0}
  \\
  KG-KR 
  & 0.244\textsubscript{2} & 0.275\textsubscript{1} 
  & 0.266\textsubscript{5} & \underline{0.247\textsubscript{1}} 
  & 0.278\textsubscript{2} & 0.276\textsubscript{2} 
  & 0.189\textsubscript{7} & 0.215\textsubscript{7} 
  & 0.311\textsubscript{1} & 0.278\textsubscript{0}
  \\
  KG-KE-KR 
  & 0.245\textsubscript{1} & 0.278\textsubscript{4} 
  & 0.267\textsubscript{3} & 0.246\textsubscript{2} 
  & \underline{0.285\textsubscript{9}} & \underline{0.279\textsubscript{4}} 
  & 0.194\textsubscript{4} & 0.220\textsubscript{2} 
  & \underline{0.314\textsubscript{0}} & \underline{0.280\textsubscript{0}}
  \\
  KG-KE-KR-M 
  & \textbf{0.257\textsubscript{2}} & \textbf{0.284\textsubscript{3}} 
  & \textbf{0.272\textsubscript{3}} & \textbf{0.250\textsubscript{2}} 
  & \textbf{0.289\textsubscript{4}} & \textbf{0.286\textsubscript{4}} 
  & \underline{0.202\textsubscript{6}} & \underline{0.223\textsubscript{3}} 
  & \textbf{0.317\textsubscript{0}} & \textbf{0.282\textsubscript{0}}
  \\
 \hline
\end{tabular}
\caption{Total keyphrase prediction results on all testing datasets. The best results are bold and the second best results are underlined. The subscripts are corresponding standard deviations for neural-based models (e.g. 0.257\textsubscript{2} means 0.257$\pm$0.002). The `\textsuperscript{*}' indicates our implementations based on \newcite{luong2015effective} attention and \newcite{see2017get} copying. The implementations of our proposed models are based on ``CopyRNN\textsuperscript{*}''.}
\label{Table:overall_keyphrase_results_kp20k_macro}
\end{table*}

\begin{table}[t]
\centering
\resizebox{\columnwidth}{!}{
\begin{tabular}{c| c| c| c| c| c}
  \hline
  \hline
  \textbf{Model} & \textbf{Inspec} & \textbf{Krapivin} & \textbf{NUS} & \textbf{SemEval} & \textbf{KP20k}
  \\
  \hline
  \hline
  TF-IDF   & 0.141 & 0.069 & 0.069 & 0.043 & 0.064\\
 TextRank  & 0.158 & 0.110 & 0.094 & 0.062 & 0.110\\
  Maui     & 0.024 & 0.162 & 0.161 & 0.012 & 0.196\\
  \hline
  CorrRNN\textsuperscript{*}  
  & 0.172\textsubscript{6} & 0.217\textsubscript{6} 
  & 0.212\textsubscript{4} & 0.125\textsubscript{3} & 0.268\textsubscript{3}\\
  CopyRNN\textsuperscript{*}  
  & 0.195\textsubscript{7} & \underline{0.229}\textsubscript{3} 
  & 0.216\textsubscript{8} & 0.120\textsubscript{8} & 0.285\textsubscript{1}\\
  \hline
  KG-KE 
  & \underline{0.197\textsubscript{3}} & 0.225\textsubscript{3} 
  & 0.219\textsubscript{3} & \textbf{0.135\textsubscript{5}} & 0.287\textsubscript{0}\\
  KG-KR 
  & 0.190\textsubscript{1} & 0.228\textsubscript{5} 
  & 0.222\textsubscript{7} & 0.120\textsubscript{6} & 0.293\textsubscript{0}\\
  KG-KE-KR 
  & 0.191\textsubscript{3} & \underline{0.229}\textsubscript{3} 
  & \underline{0.224\textsubscript{5}} & 0.127\textsubscript{5} & \underline{0.295\textsubscript{1}}\\
  KG-KE-KR-M 
  & \textbf{0.201\textsubscript{2}} & \textbf{0.234}\textsubscript{2}
  & \textbf{0.234\textsubscript{6}} & \underline{0.131\textsubscript{4}} & \textbf{0.299\textsubscript{0}}\\
  \hline
\end{tabular}
}
\caption{MAP@10 scores of total keyphrase predictions. The best results are bold and the second best results are underlined. The meanings of the subscripts and the `\textsuperscript{*}' are the same as in Table~\ref{Table:overall_keyphrase_results_kp20k_macro}.}
\label{table:map10 measure for total keyphrases}
\end{table}

\subsection{Implementation Details}
We apply similar preprocessing procedures with~\newcite{Meng2017dkg} including lowercasing, tokenizing and replacing digits with $\langle digit \rangle$ symbol. The title and the abstract of each article are concatenated as the source text input. We use the \textbf{KP20k} training dataset as the retrieval corpus. The implementations of our models are based on the OpenNMT system~\cite{klein2017opennmt}. The encoders, the decoder, and the scorer have the same vocabulary $V$ with 50,000 tokens. The multi-task learning model and the scorer are trained separately.

The embedding dimension $d_e$ and the hidden size $d$ are set to 100 and 300 respectively. The initial state of the decoder GRU cell (i.e., $\mathbf{h}_0$) is set to $[\overrightarrow{\mathbf{u}}_{L_{\mathbf{x}}}; \overleftarrow{\mathbf{u}}_1]$. The other GRU cells are set to zero. The retrieval number $K$ is set to 3 after evaluating the retrieved keyphrases on the evaluation dataset. When concatenating the retrieved keyphrases together as an external knowledge input, we use `;' as the separator among them. During training, all the trainable parameters including the embeddings are randomly initialized with uniform distribution in [-0.1, 0.1]. We engage Adam~\cite{Kingma2014Adam} as the optimizer with positive extraction loss weight $w$=9.0, batch size=64, dropout rate=0.1, max gradient norm=1.0, initial learning rate=0.001. 
The training is early stopped when the validation perplexity stops dropping for several continuous evaluations. While testing, the beam search depth, and beam size are set as 6 and 200 correspondingly. The extraction threshold $\epsilon$ is set to 0.7 after evaluating the extracted keyphrases on the evaluation dataset. Notably, the stemmer is not applied to the gold keyphrases of \textbf{SemEval} testing dataset since they have already been stemmed. We do not remove any single-word predictions for \textbf{KP20k} but only keep one single-word prediction for other testing datasets. The averaged results of three different random seed are reported\footnote{Our code is available at https://github.com/Chen-Wang-CUHK/KG-KE-KR-M}.

\section{Results and Analysis}
\subsection{Total Keyphrase Prediction} \label{section:total_keyphrase_prediction}
Unlike the previous works which only separately analyze the present and absent keyphrase prediction ability, we also compare the whole keyphrase prediction ability regardless of the presence or absence of keyphrases, which is more reasonable in real applications. We show the $\text{F}_1$ scores at the top 5 and 10 predictions on Table~\ref{Table:overall_keyphrase_results_kp20k_macro}.

This table displays our KG-KE-KR-M method consistently outperforms the state-of-the-art models CopyRNN and CorrRNN demonstrating the effectiveness of our method. Moreover, we also observe that the KG-KE model exceeds CopyRNN and CorrRNN on most datasets, which indicates the strength of our combination with the extractive model. Besides, we also see the KG-KR model perform comparably or better than the baselines, suggesting the effective guidance ability of the retrieved keyphrases. In addition, after combining these two ideas, the KG-KE-KR model surpasses both or one of KG-KE and KG-KR on all datasets, which shows the effectiveness of the combination with extraction model and the retrieved keyphrases again. Finally, the performance gap between KG-KE-KR and KG-KE-KR-M implies the power of our merging module. For mean average precision (MAP) metric which considers prediction orders, we obtain similar conclusions as shown in Table~\ref{table:map10 measure for total keyphrases}.

\begin{figure}[t]
\centering
\includegraphics[width=0.9\columnwidth]{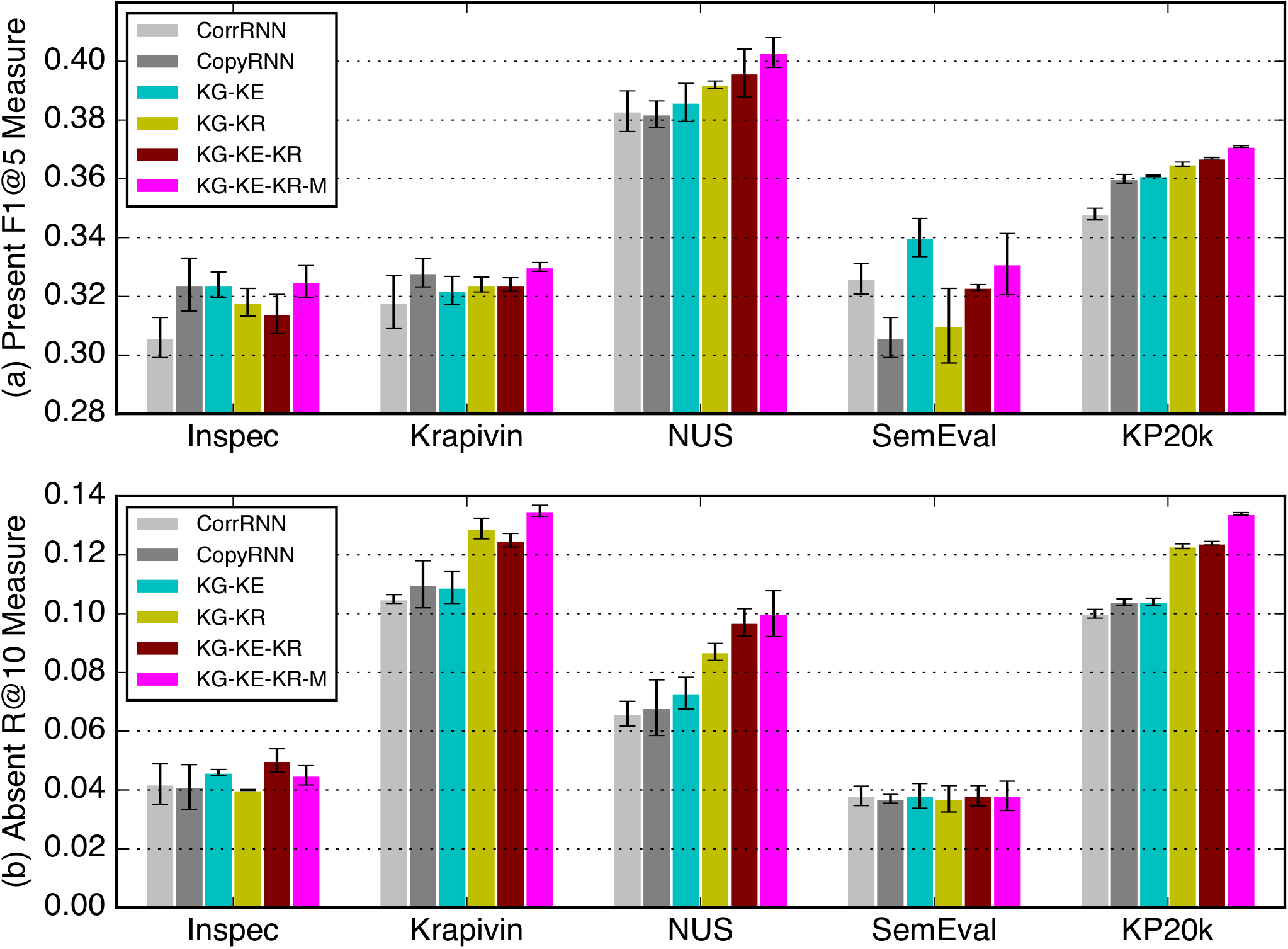}
\caption{The present and absent keyphrase prediction performance of all neural-based methods.}
\label{figure: present_absent_hist}
\end{figure}

\subsection{Present and Absent Keyphrase Prediction}
In this section, we analyze the performance of present and absent keyphrase prediction. Only the present (absent) predictions and gold present (absent) keyphrases are preserved for the corresponding evaluation. We use $\text{F}_1$@5 metric for present predictions and R@10 for absent predictions. Since the neural-based baselines are the state-of-the-art models, we focus on the comparison with them in this section. The results are depicted on Figure~\ref{figure: present_absent_hist}.

The main observations are similar to the conclusions of total keyphrase prediction. Besides, we also note that after incorporating retrieved keyphrases, KG-KR model achieves substantial improvement gains over baselines on absent keyphrase prediction on \textbf{Krapivin}, \textbf{NUS}, and \textbf{KP20k}. These results demonstrate that the retrieved keyphrases indeed help the model to understand the main topics of the given document since generating absent keyphrase is an abstractive process and requires more powerful text understanding abilities. 
We notice that the KG-KE-KR-M method does not outperform the KG-KE-KR model on absent keyphrase prediction on \textbf{Inspec} dataset. One potential reason is that the merging module only merges two sources for absent keyphrases (i.e., the generated and retrieved keyphrases) instead of three sources like the present keyphrases do. Hence, the improvement for the absent keyphrases from the merging module is less stable than that for the present keyphrases.
Moreover, we find that after combining with the extraction model, the KG-KE model achieves a huge improvement gain over CopyRNN on present keyphrase prediction on \textbf{SemEval} dataset, which manifests such a combination can improve the keyphrase extraction ability of the generative model.

\begin{table}[t]
\centering
\resizebox{\columnwidth}{!}{
\begin{tabular}{c| c| c| c}
  \hline
  \hline
 Candidate Sources & Total $\text{F}_1@10$ & Present $\text{F}_1@5$ & Absent $\text{R}@10$
  \\
  \hline
  \hline
 $\mathbf{gk, ek, rk}$ & \textbf{0.250$\pm$0.002} & \textbf{0.330$\pm$0.002} & \textbf{0.172$\pm$0.002}\\
  \hline
 $\mathbf{gk, ek}$ & 0.249$\pm$0.003 & 0.328$\pm$0.003 & 0.154$\pm$0.002\\
 \hline
 $\mathbf{gk, rk}$ & 0.249$\pm$0.002 & 0.329$\pm$0.002 & 0.172$\pm$0.002\\
 \hline
 $\mathbf{gk}$ & 0.248$\pm$0.003 & 0.327$\pm$0.003 & 0.154$\pm$0.002\\
 \hline
 $\mathbf{gk}$, no merging & 0.246$\pm$0.002 & 0.324$\pm$0.002 & 0.158$\pm$0.002\\
  \hline
 $\mathbf{ek}$, no merging & 0.152$\pm$0.005 & 0.226$\pm$0.010 & N/A\\
 \hline
 $\mathbf{rk}$, no merging & 0.093$\pm$0.000 & 0.121$\pm$0.000 & 0.107$\pm$0.000\\
  \hline
\end{tabular}
}
\caption{Ablation study of the candidate sources of Algorithm~\ref{alg:merging} on \textbf{Krapivin} dataset. ``no merging'' means we do not use the merging algorithm.}
\label{table:ablation study of candidate sources}
\end{table}

\begin{figure*}[t]
\centering
\includegraphics[width=\textwidth]{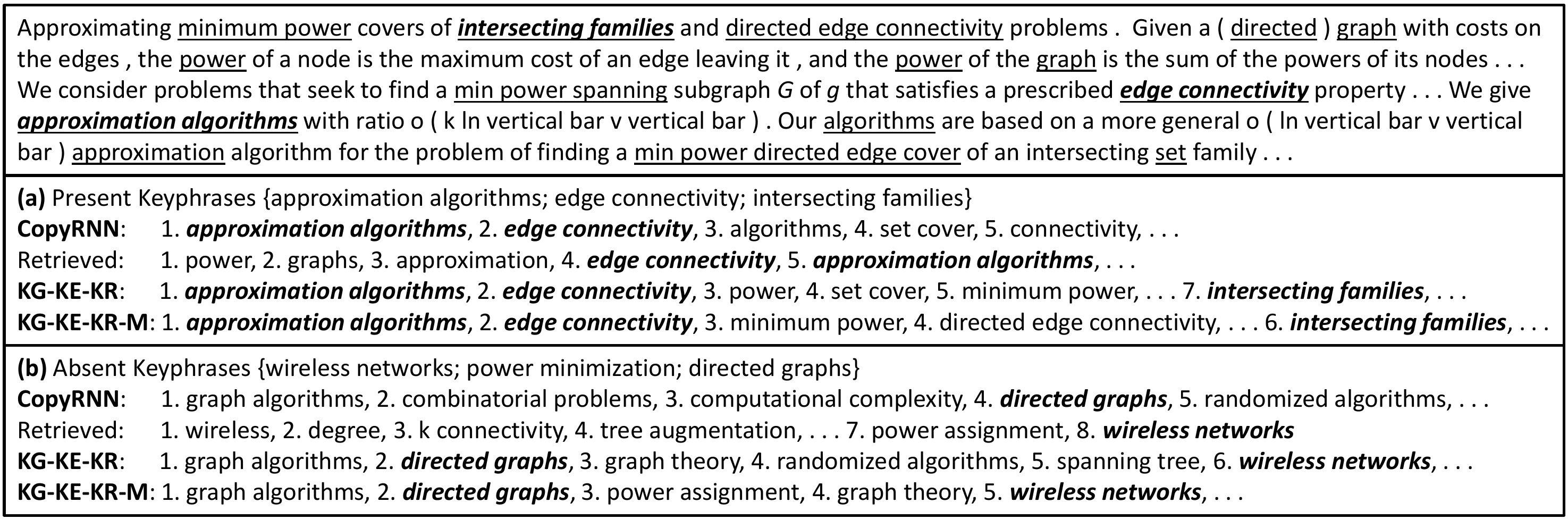}
\caption{A keyphrase prediction example of CopyRNN, KG-KE-KR, and KG-KE-KR-M. ``Retrieved'' is the retrieved keyphrases. The extracted keyphrases by the extractor of KG-KE-KR are underlined in the source text. Top 10 present and absent predictions are compared and some incorrect predictions are omitted for simplicity. The correct predictions are bold and italic.}
\label{figure: case_study_final}
\end{figure*}

\subsection{Ablation Study on Merging Module}
We also conduct in-depth ablation studies on our merging module. The objectives of these ablation studies are to (1) evaluate the effects of different candidate sources (i.e., what kinds of candidates are merged), and (2) analyze the effects of different final importance score calculating methods.

Concerning candidate sources, we show the ablation study results on Table~\ref{table:ablation study of candidate sources}. When comparing ``$\mathbf{gk}$'' with ``$\mathbf{gk}$, no merging'', we can see that the merging algorithm improves the performance of total and present keyphrase predictions, but it degrades the performance of absent keyphrase prediction. These results indicate the trained \textit{scorer} performs better on scoring present keyphrases than scoring absent keyphrases.
One possible reason is that scoring absent keyphrases requires a stronger text understanding ability than scoring present keyphrases. However, as shown in the row of ``$\mathbf{gk, rk}$'' on Table\ref{table:ablation study of candidate sources}, this problem can be solved by incorporating the retrieved keyphrases which provide external information to this module. Besides absent keyphrase prediction, it is observed that the retrieved keyphrases can also benefit the present keyphrase prediction. For the extracted keyphrases, as shown in the ``$\mathbf{gk, ek}$'' row, they only improve the present keyphrase prediction ability and do not affect absent keyphrases as we anticipated.

Regarding the scoring method, we further explore the effects of not using or only using the \textit{scorer} in Algorithm~\ref{alg:merging}. We show the results on Table~\ref{table:ablation study of scoring methods}. From this table, we note that after removing the \textit{scorer} (i.e., ``Only $\mathbf{gs, es, rs}$''), both present and absent keyphrase prediction performance become worse, which demonstrates the effectiveness of the combination with the \textit{scorer}. Moreover, if we totally ignore the previously obtained retrieval, extraction and generation scores, and only use the \textit{scorer} to predict the final keyphrase importance score (i.e., ``Only \textit{scorer}''), we find the performance decreases dramatically, which indicates the indispensability of the previously obtained retrieval, extraction, and generation scores.

\begin{table}[t]
\centering
\resizebox{\columnwidth}{!}{
\begin{tabular}{c| c| c| c}
  \hline
  \hline
 Scoring Method & Total $\text{F}_1@10$ & Present $\text{F}_1@5$ & Absent $\text{R}@10$
  \\
  \hline
  \hline
  Combined & \textbf{0.250$\pm$0.002} & \textbf{0.330$\pm$0.002} & \textbf{0.172$\pm$0.002}\\
  \hline
 Only $\mathbf{gs, es, rs}$ & 0.248$\pm$0.003 & 0.325$\pm$0.003 & 0.166$\pm$0.003\\
 \hline
 Only \textit{scorer} & 0.210$\pm$0.005 & 0.291$\pm$0.006 & 0.106$\pm$0.005\\
  \hline
\end{tabular}
}
\caption{Ablation study of the scoring method of Algorithm~\ref{alg:merging} on \textbf{Krapivin} dataset. ``Only $\mathbf{gs, es, rs}$'' means we do not use the \textit{scorer}. ``Only \textit{scorer}'' represents we directly use the scores predicted by the \textit{scorer} as the final importance scores.}
\label{table:ablation study of scoring methods}
\end{table}

\subsection{Case Study}
To illustrate the advantages of our proposed methods, we show an example of the present and absent keyphrase predictions in Figure~\ref{figure: case_study_final}. For fairness, we only compare with CopyRNN since our models are based on its implementation. From the results of the present keyphrase prediction, we find the extractor of the KG-KE-KR model successfully extracts all the present keyphrases from the source text, which shows the power of the extractor. With the help of the copy probability rescaling from the extractor, the KG-KE-KR model correctly predicts the keyphrase ``intersecting families'' which is not successfully predicted by CopyRNN and retrieved by the retriever. Moreover, by merging the extracted keyphrases into the final predictions, the KG-KE-KR-M model assigns a higher rank to this keyphrase (i.e., from 7 to 6). As for absent keyphrase prediction, we note that KG-KE-KR successfully predicts the keyphrase ``wireless networks'' while CopyRNN fails. Since the retriever successfully retrieves this absent keyphrase, it shows that the retrieved keyphrases can provide effective external guidance for the generation process. Furthermore, the KG-KE-KR-M method assigns a higher rank to this keyphrase after merging the retrieved keyphrases into the final predictions (i.e., from 6 to 5). The overall results demonstrate the effectiveness of our proposed methods.

\section{Conclusion and Future Work}
In this paper, we propose a novel integrated approach for keyphrase generation. First, an end-to-end multi-task learning framework is introduced, which not only combines the keyphrase extraction and generation but also leverages the retrieved keyphrases from similar documents to guide the generation process. Furthermore, we introduce a neural-based merging algorithm to merge the candidates from three different components. Comprehensive empirical studies demonstrate the effectiveness of our approach. One interesting future work is to incorporate the similar documents themselves into keyphrase generation. 

\section*{Acknowledgments}
The work described in this paper was partially supported by the Research Grants Council of the Hong Kong Special Administrative Region, China (No.~CUHK~14208815 of the General Research Fund) and Meitu~(No.~7010445). We would like to thank Jiani Zhang for her comments.

\bibliographystyle{acl_natbib}
\bibliography{naaclhlt2019}

\end{document}